\documentclass[10pt,twocolumn,letterpaper]{article}

\usepackage{cvpr}
\usepackage{times}
\usepackage{epsfig}
\usepackage{graphicx}
\usepackage{amsmath}
\usepackage{amssymb}
\usepackage[normalem]{ulem}
\usepackage{subcaption}
\usepackage{arydshln}
\usepackage{xcolor}
\usepackage{etoolbox}
\usepackage[font={small}]{caption}
\usepackage[nocompress]{cite} %

\usepackage[inline]{enumitem}
\usepackage{enumitem}%

\usepackage[pagebackref=true,breaklinks=true,colorlinks,bookmarks=false]{hyperref}

\cvprfinalcopy %

\def\cvprPaperID{2392} %

\definecolor{brown}{rgb}{0.59, 0.29, 0.0}
\definecolor{darkgreen}{rgb}{0, 0.569, 0}
\definecolor{darkyellow}{rgb}{0.9, 0.9, 0.2}

\ifcvprfinal\fi

\begin{document}

\providetoggle{showcomments}									%
\settoggle{showcomments}{true} 								%

\providetoggle{longversion}										%
\settoggle{longversion}{true} 								%

\newcommand{\para}[1]{\vspace{3pt} \noindent \textbf{#1}}
\newcommand{\att}[1]{\textcolor{black}{{#1}}} %

\newcommand{\coco}{COCO}
\newcommand{\ilsvrc}{ILSVRC}
\newcommand{\imagenet}{ImageNet}
\newcommand{\taskname}{object class labelling}

\title{Fast Object Class Labelling via Speech}

\author{Michael Gygli\\
{\tt\small gyglim@google.com} \\
\and
Vittorio Ferrari\\
{\tt\small vittoferrari@google.com}
}

\maketitle

\begin{abstract}
Object class labelling is the task of annotating images with labels on the presence or absence of objects from a given %
class vocabulary.
Simply asking one yes/no question per class, however, has a cost that is linear in the vocabulary size and is thus inefficient for large vocabularies.
Modern approaches rely on a hierarchical organization of the vocabulary to reduce annotation time, but remain expensive (several minutes per image for the 200 classes in \ilsvrc{}).
Instead, we propose a new interface where classes are annotated via speech.
Speaking is fast and allows for direct access to the class name, without searching through a list or hierarchy.
As additional advantages, annotators can simultaneously speak and scan the image for objects, the interface can be kept extremely simple, and using it requires less mouse movement. %
As annotators using our interface should only say words from a given class vocabulary, we propose a dedicated task that trains them to do so.
Through experiments on COCO and ILSVRC, we show our method yields high-quality annotations at
\att{$2.3{\times}-14.9{\times}$} less annotation time than existing methods.

\end{abstract}

\section{Introduction}
\begin{figure}[t]
	\centering\includegraphics[trim={0 0 0 2cm},clip,width=1\linewidth]{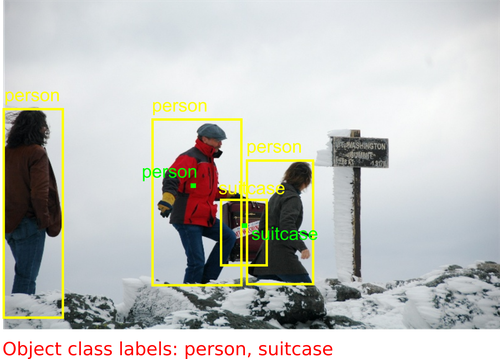}
   \caption{Illustration of common stages of image annotation: typically annotators first provide object class labels at the image-level~\cite{deng09cvpr,kuznetsova18arxiv} (\textcolor{red}{red}), sometimes associated to a specific object via a click as in~\cite{lin14eccv} and our approach (\textcolor{green}{green}).
	Following stages then annotate the spatial extent of objects, \eg with bounding boxes or segmentations (\textcolor{yellow}{yellow}).}
	\label{fig:task_illustration}
\end{figure}

Deep neural networks need millions of training examples to obtain high performance. Therefore, large and diverse datasets such as \ilsvrc{}~\cite{deng09cvpr}, COCO~\cite{lin14eccv} or Open Images~\cite{kuznetsova18arxiv} lie at the heart of the breakthrough and ongoing advances in visual recognition.

Datasets for recognition are typically annotated in two stages~\cite{deng09cvpr,kuznetsova18arxiv,lin14eccv,su12aaai} (Fig.~\ref{fig:task_illustration}):
(i) determining the presence or absence of object classes in each image, and
(ii) providing bounding boxes or segmentation masks for all classes present.
Our work focuses on the former, which we call \textit{\taskname{}}.
As marking a class as present requires finding at least one object of that class, we also ask annotators to click on it (as also done for the COCO dataset~\cite{lin14eccv}).
This task is not only natural, it also helps the subsequent annotation stages~\cite{lin14eccv}, and can be used as input to weakly-supervised methods~\cite{bearman16eccv,mettes16eccv,manen17iccv,papadopoulos17cvpr}.

Object class labelling has traditionally been time-consuming for annotators.
A na\"ive approach is to ask a separate yes/no question for each class of a given vocabulary.
Such a protocol is rooted on the vocabulary, not the image content. It scales linearly in the size of the vocabulary, even when only few of the classes are present in the image (which is the typical case).
Thus, it becomes very inefficient when the vocabulary is large.
Let's take the \ilsvrc{} dataset as an example:
getting labels for the 200 object classes in the vocabulary would take close to \att{6 minutes per image~\cite{krishna16chi}},
despite each image containing only 1.6 classes on average.
Previous methods have attempted to improve on this by using a hierarchical representation of the class vocabulary to quickly reject certain groups of labels~\cite{lin14eccv,deng14chi}.
This reduces the annotation complexity to sub-linear in the vocabulary size.
But even with these sophisticated methods, \taskname{} remains time consuming.
Using the hierarchical method of~\cite{deng14chi} to label the 200 classes of~\ilsvrc{} still takes 3 minutes per image~\cite{russakovsky15cvpr}.
The COCO dataset has fewer classes (80) and was labelled using the more efficient hierarchical method of~\cite{lin14eccv}. Even so, it still took half a minute per image.

In this paper, we improve upon these approaches by using speech as an input modality.
Given an image, annotators scan it for objects and mark one per class by clicking on it and saying its name.
This task is rooted on the image content and naturally scales with the number of object classes in the image.
Using speech has several advantages:
(i) it allows for direct access to the class name via simply \textit{saying it}, rather than requiring a hierarchical search.
(ii) it does not require the experiment designer to construct a natural, intuitive hierarchy, which becomes difficult as the class vocabulary grows~\cite{russakovsky15ijcv}.
(iii) combining speaking with pointing is natural and efficient:
when using multimodal interfaces, people naturally choose to point for providing spatial information and to speak for semantic information~\cite{oviatt03book}. Also, these two tasks can be done concurrently~\cite{kahneman73attention, oviatt03book}.
(iv) As the class label is provided via speech, the task requires less mouse movement and the interface becomes extremely simple
(no need to move back and forth between the image and the class hierarchy representation).
(v) Finally, speaking is fast, \eg people can say 150 words per minute when describing images~\cite{vaidyanathan18acl}.
In comparison, people normally type at 30-100 words per minute~\cite{karat99sigchi,clarkson05chi}.
Thanks to the above points, our interface is more time efficient than hierarchical methods.

Using speech as an input modality, however, poses certain challenges.
In order to reliably transcribe speech to text,
several technical challenges need to be tackled, such as segmenting the speech and obtaining high-accuracy transcriptions.
Furthermore, as speech is free-form in nature, annotators need to be trained to know the class vocabulary%
\iftoggle{longversion}{ to be annotated}, in order to not label other objects or forget to annotate some classes.
We show how to tackle these challenges and design an annotation interface that allows for fast and accurate \taskname{}.

In our extensive experiments we:
\begin{itemize}[noitemsep,topsep=0pt]
    \item Show that speech provides a fast way for \taskname{}: \att{2.3$\times$} faster on the \coco{} dataset~\cite{lin14eccv} than the hierarchical approach of~\cite{lin14eccv}, and \att{14.9$\times$} faster than~\cite{deng14chi} on \ilsvrc{}~\cite{russakovsky15ijcv}.
    \item Demonstrate the ability of our method to scale to large vocabularies.
    \item Show that our interface enables to carry out the task with $3\times$ shorter mouse paths than~\cite{lin14eccv}.
	\item Show that through our training task annotators learn to use the provided vocabulary for naming objects with high fidelity.
    \item Analyse the accuracy of models for automatic speech recognition (ASR) and show that it supports deriving high-quality annotations from speech.
\end{itemize}

\section{Related Work}
Using speech as an input modality has a long history~\cite{bolt80siggraph} and is recently emerging as a research direction in Computer Vision~\cite{dai16thesis, vasudevan17cvpr, vaidyanathan18acl, harwath18eccv}.
To the best of our knowledge, however, our paper is the first to show that speech allows for more efficient \taskname{} than the prevailing hierarchical approaches~\cite{lin14eccv,deng14chi}.
We now discuss previous works in the areas of leveraging speech, efficient image annotation and learning from point supervision.

\para{Leveraging speech inputs.}
To point and speak is an efficient and natural way of human communication.
Hence, this approach was quickly adopted when designing computer interfaces: as early as 1980, Bolt~\cite{bolt80siggraph} investigates using speech and gestures for manipulating shapes.
Most previous works in this space analyse what users choose when offered different input modalities~\cite{hauptmann89sigchi,oviatt96sigchi,oviatt97integration,oviatt03book},
while only a few approaches focus on the added efficiency of using speech.
The most notable such work is~\cite{pausch91vio}, which measures the time needed to create a drawing in MacDraw.
They compare using the tool as is, which involves selecting commands via the menu hierarchy, to using voice commands. They show that using speech gives an average speedup of 21\% and mention this is a ``lower bound'', as the tool was not designed with speech in mind.

In Computer Vision, Vasudevan~\etal~\cite{vasudevan17cvpr} detect objects given spoken referring expressions,
while Harwath~\etal.~\cite{harwath18eccv} learn an embedding from spoken image-caption pairs. Their approach obtains promising first results, but still performs inferior to learning on top of textual captions obtained from Google's automatic speech recognition.
Damen~\etal~\cite{damen18eccv} annotates the EPIC-KITCHENS dataset based on spoken free-form narratives, which cover only some of the objects present in the image.
Moreover, these narratives are transcribed {\em manually}, and then object class labels are derived from transcribed nouns, again manually.
Instead, our approach is fully automatic and we exhaustively label all objects from a given vocabulary.
Finally, more closely related to our work, Vaidyanathan~\etal\cite{vaidyanathan18acl} re-annotated a subset of \coco{} with spoken scene descriptions and human gaze.
While efficient, free-form scene descriptions are more noisy when used for \taskname{}, as annotators might refer to objects with ambiguous names, mention nouns that do not correspond to objects shown in the image~\cite{vaidyanathan18acl}, or there might be inconsistencies in naming the same object classes across different annotators.
Our approach avoids the additional complexities of parsing free-form sentences to extract object names and gaze data to extract object locations.

\para{Sub-linear annotation schemes.}
The na\"ive approach to annotating the presence of object classes grows linearly with the
size of the vocabulary (one binary present/absent question per class).
The idea behind sub-linear schemes is to group the classes into meaningful super-classes, such that several of them can be ruled out at once. If a super-class (\eg animals) is not present in the image, then one can skip the questions for all its subclasses (cat, dog, \etc).
This grouping of classes can have multiple levels.
The annotation schemes behind \coco{}~\cite{lin14eccv} and ~\ilsvrc{}~\cite{deng14chi,russakovsky15ijcv} datasets both fall into this category,
but they differ in how they define and use the hierarchy.

~\ilsvrc{}~\cite{russakovsky15ijcv} was annotated using a series of hierarchical questions~\cite{deng14chi}.
For each image, 17 top-level questions were asked (\eg ``Is there a living organism?''). For groups that are present, more specific questions are asked subsequently, such as ``Is there a mammal?'', ``Is there a dog?'',~\etc.
The sequence of questions for an image is chosen dynamically, such that the they allow to eliminate the maximal number of labels at each step~\cite{deng14chi}.
This approach, however, involves repeated visual search, in contrast to ours, which is guided by the annotator scanning the image for objects, done only once.
Overall, this scheme takes close to 3 minutes per image~\cite{russakovsky15cvpr} for annotating the 200 classes of~\ilsvrc{}.
On top of that, constructing such a hierarchy is not trivial and influences the final results~\cite{russakovsky15ijcv}.

In the protocol used to create \coco{}~\cite{lin14eccv}, annotators are asked to mark one object for each class present in an image
by choosing its symbol from a two-level hierarchy and dragging it onto the object (Fig.~\ref{fig:hierarchical_interface}).
While this allows to take the image, rather than the questions as the root of the labelling task, it requires repeatedly searching for the right class in the hierarchy, which induces significant time cost.
In our interface, such an explicit class search is not needed, which speeds up the annotation process.

Rather than using a hierarchy, Open Images~\cite{kuznetsova18arxiv} uses an image classifier to create a shortlist of object classes likely to be present, which are then verified by annotators using binary questions.
The shortlist is generated using a pre-defined threshold on the classifier scores. Thus, this approach trades off completeness for speed.
In practice,~\cite{kuznetsova18arxiv} asks annotators to verify 10 out of 600 classes, but report a rather low recall of 59\%, despite disregarding ``difficult'' objects in evaluation.

\para{Point supervision.}
The output of our annotation interface is a list of all classes present in the image with a point on one object for each.
This kind of labelling is efficient and provides useful supervision for several image~\cite{papadopoulos17cvpr, bearman16eccv, laradji18arxiv} and video~\cite{mettes16eccv,manen17iccv} object localization tasks. %
In particular,~\cite{papadopoulos17cvpr, bearman16eccv, manen17iccv} show that for their task, point clicks deliver better models than other alternatives when given the same annotation budget.

\begin{figure}[t]
		\centering
		\includegraphics[width=1\linewidth]{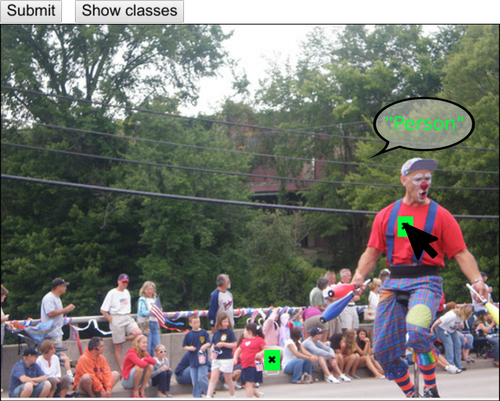}
		\caption{\textbf{Our interface.} Given an image the annotator is asked to click on one object per class and say its name.
		To aid memory, we additional allow to review the class vocabulary through the ``Show classes'' button.
		}
	\label{fig:interface}
\end{figure}

\section{Speech-based annotation}
We now describe our annotation task, which produces a series of time-stamped click positions $\{p_i\}$ and an audio recording for each image (Sec.~\ref{sec:main_task}).
From this, we obtain object class labels by associating audio segments to clicks and then transcribing the audio (Sec.~\ref{sec:postprocessing}).
Before annotators can proceed to the main task, we require them to pass a training stage. This helps them memorise the class vocabulary and get confident with using the interface (Sec.~\ref{sec:training}).

\subsection{Annotation task}
\label{sec:main_task}
First, annotators are presented with the class vocabulary and instructed to memorise it.
Then, they are asked to label images with object classes from the vocabulary, by scanning the image and saying the names of the different classes they see.
Hence, this is a simple visual search task that does not require any context switching.
While we are primarily interested in object class labels, we ask annotators to click on one object for each class, as the task naturally involves finding objects anyway. Also, this brings valuable additional information, and matches the COCO protocol, allowing for direct comparisons (Sec.~\ref{sec:experiments_baseline}).
Fig.~\ref{fig:interface} shows the interface with an example image.

To help annotators restrict the labels they provide to the predefined vocabulary, we allow them to review it using a button that shows all class names including their symbols.
\begin{figure}[t]
\centering
   \begin{subfigure}[b]{1\linewidth}
   \centering
		\includegraphics[width=1.0\linewidth]{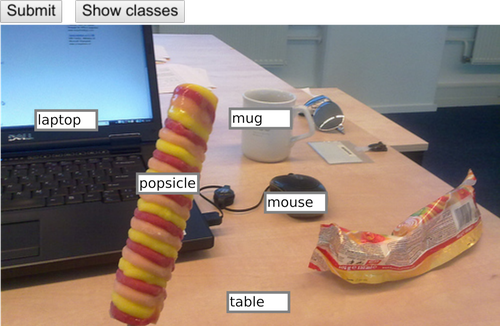}
        \caption{}
        \label{fig:qualification_task}
        \vspace{0.2cm}
    \end{subfigure}
   \begin{subfigure}[b]{1\linewidth}
		\includegraphics[width=1\linewidth]{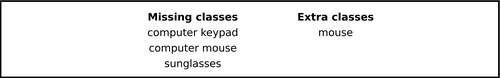}		
        \caption{}
        \label{fig:qualification_feedback}
    \end{subfigure}
		\caption{\textbf{Training process}. \ref{fig:qualification_task} shows the training task: marking an object per class with a click and saying and writing its name.
		\ref{fig:qualification_feedback} shows the feedback provided after each image.
		}
	\label{fig:qualification}
\end{figure}

\subsection{Temporal segmentation and transcription}
\label{sec:postprocessing}

In order to assign class names to clicks, we need to transcribe the audio and temporally align the transcriptions.
To obtain transcriptions and their start and end time we rely on Google's automatic speech recognition API\footnote{\url{https://cloud.google.com/speech-to-text/}}.
While it would be possible to first transcribe the full audio recording and then match the transcriptions to clicks, we found that the temporal segmentation of transcriptions is error-prone.
Hence, we opt to first segment the audio recording based on the clicks' timestamps and then transcribe these segments.

\para{Temporal segmentation of the recording.}
We create an object annotation $o_i$ for each click at position $p_i$ and time $t_i$.
For each object annotation we create an audio segment $[t_i - \delta, t_{i+1}]$,
~\ie an interval ranging from shortly before the current click to the next click.
Finally, we transcribe these audio segments and assign the result to their corresponding object annotations $o_i$.
Empirically, using a small validation set, we found that $\delta=0.5$s performs best,
as people often start speaking slightly before clicking on the object~\cite{oviatt03book}.

\para{Transcribing the object class name.}
The speech transcription provides a ranked list of alternatives.
To find the most likely class in the vocabulary we use the following algorithm:
(i) if one or more transcriptions match a class in the vocabulary, we use the highest ranking;
ii) in the rare case that none matches, we represent the vocabulary and all the transcriptions using word2vec~\cite{mikolov13arxiv} and use the most similar class from the vocabulary,
according to their cosine similarity.
This class $c_i$ is then treated as the label of $o_i$.

\subsection{Annotator training}
\label{sec:training}
Before tackling the main task, annotators go through a training stage which provides feedback after every image and also aggregated statistics after 80 images.
If they meet our accuracy targets, they can proceed to the main task. If they fail, they can repeat the training until they succeed.

\para{Purpose of training.}
Training helps annotators to get confident with an interface and allows to ensure they correctly solve the task and provide high-quality labels.
As a consequence, it has become common practice~\cite{russakovsky15ijcv,su12aaai, lin14eccv, kuznetsova18arxiv, papadopoulos17cvpr}.%

While we want to annotate classes from a predefined vocabulary, speech is naturally free-form.
In our initial experiments we found that annotators produced lower recall compared to an interface which displays an explicit list of classes due to this discrepancy.
Hence, we designed our training task to ensure annotators memorise the vocabulary and use the correct object names.
Indeed, after training annotators with this process they rarely use object names that are not in the vocabulary and obtain a high recall, comparable to~\cite{lin14eccv} (Sec.~\ref{sec:experiments_coco} \& ~\ref{sec:detailed_experiments}).

\para{Training procedure.}
The training task is similar to the main task, but we additionally require annotators to type the words they say (Fig.~\ref{fig:qualification_task}).
This allows to measure transcription accuracy and dissect different sources of error in the final class labelling (Sec.~\ref{sec:detailed_experiments}).
After each image we provide immediate feedback listing their mistakes, by comparing their answers against a pre-annotated ground truth.
This helps annotators memorise the class vocabulary and learn to spot all object classes (Fig.~\ref{fig:qualification_feedback}).
We base this feedback on the written words, rather than the transcribed audio, for technical simplicity.

\para{Passing requirements.}
At the beginning of training, annotators are given targets on the minimum recall and precision they need to reach.
Annotators are required to label 80 images and are given feedback after every image, listing their errors on that image, and on how well they do overall with respect to the given targets.
If they meet the targets after labelling 80 images, they successfully pass training.
In case of failure, they are allowed to repeat the training as many times as they want.

\section{Experiments}
Here we present experiments on annotating images using our speech-based interface and the hierarchical interface of~\cite{lin14eccv}.
First, in Sec.~\ref{sec:experiments_baseline} we reimplement the interface of~\cite{lin14eccv} and compare it to the official reported results in~\cite{lin14eccv}.
Then, we compare our interface to that of~\cite{lin14eccv} on the COCO dataset, where the vocabulary has 80 classes (Sec.~\ref{sec:experiments_coco}). In Sec.~\ref{sec:experiments_imagenet} we scale up annotation to a vocabulary of 200 classes by experimenting on the \ilsvrc{} dataset.
Finally, Sec.~\ref{sec:detailed_experiments} provides additional analysis such as the transcription and click accuracy as well as response times per object.

\subsection{Hierarchical interface of~\cite{lin14eccv}}
\label{sec:experiments_baseline}
\begin{figure}[t]
	\centering\includegraphics[width=1\linewidth]{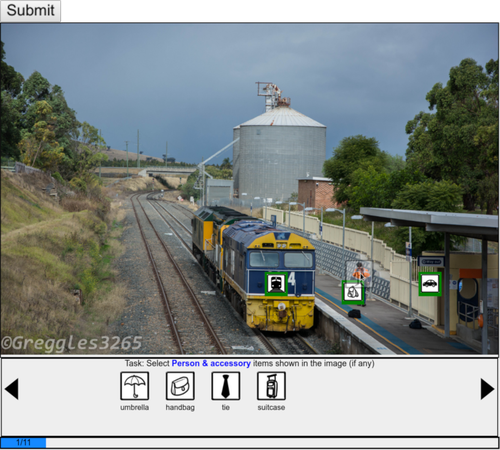}
		\caption{Our reimplementation of the hierarchical interface of~\cite{lin14eccv}.}
	\label{fig:hierarchical_interface}
\end{figure}
In the interface used for \coco~\cite{lin14eccv}, annotators are asked to mark one object for each class present in an image by choosing its symbol from a two-level hierarchy and dragging it onto the object. While~\cite{lin14eccv} provides coarse timings, we opted to re-implement their interface for fair comparison and to do a detailed analysis on how annotation time is spent (Fig.~\ref{fig:hierarchical_interface}).
First, we made five crowd workers pass a training task equivalent to that used for our interface (Sec.~\ref{sec:training}).
Then, they annotated a random subset of 300 images of the \coco{} validation set (each image was annotated by all workers).

\para{Results.}
Annotators take \att{$29.9$} seconds per image on average, well in line with the $27.4$ seconds reported in~\cite{lin14eccv}.
Hence, we can conclude that our implementation is equivalent in terms of efficiency.

Annotators have produced annotations with \att{89.3\%} precision and \att{84.7\%} recall against the ground-truth (Tab.~\ref{tab:accuracy}).
Thus, they are accurate in the labels they provide and recover most object classes. We also note that the \coco{} ground-truth itself is not free of errors, hence limiting the maximal achievable performance.
Indeed, our recall and precision are comparable to the numbers reported in~\cite{lin14eccv}.

\para{Time allocation.}
In order to better understand how annotation time is spent, we recorded mouse and keyboard events.
This allows us to estimate the time spent on searching for the right object class in the hierarchy of symbols and measure the time spent dragging the symbol.
On average, search time is \att{$14.8$}s and drag time \att{$3.4$}s per image.
Combined, these two amount to \att{$61\%$} of the total annotation time, while the rest is spent on other tasks such as visual search.
This provides a target on the time that can be saved by avoiding these two operations, as done in our interface. In the remainder of this section, we compare our speech-based approach against this annotation method.

\begin{figure}[t]
	\centering\includegraphics[width=1\linewidth]{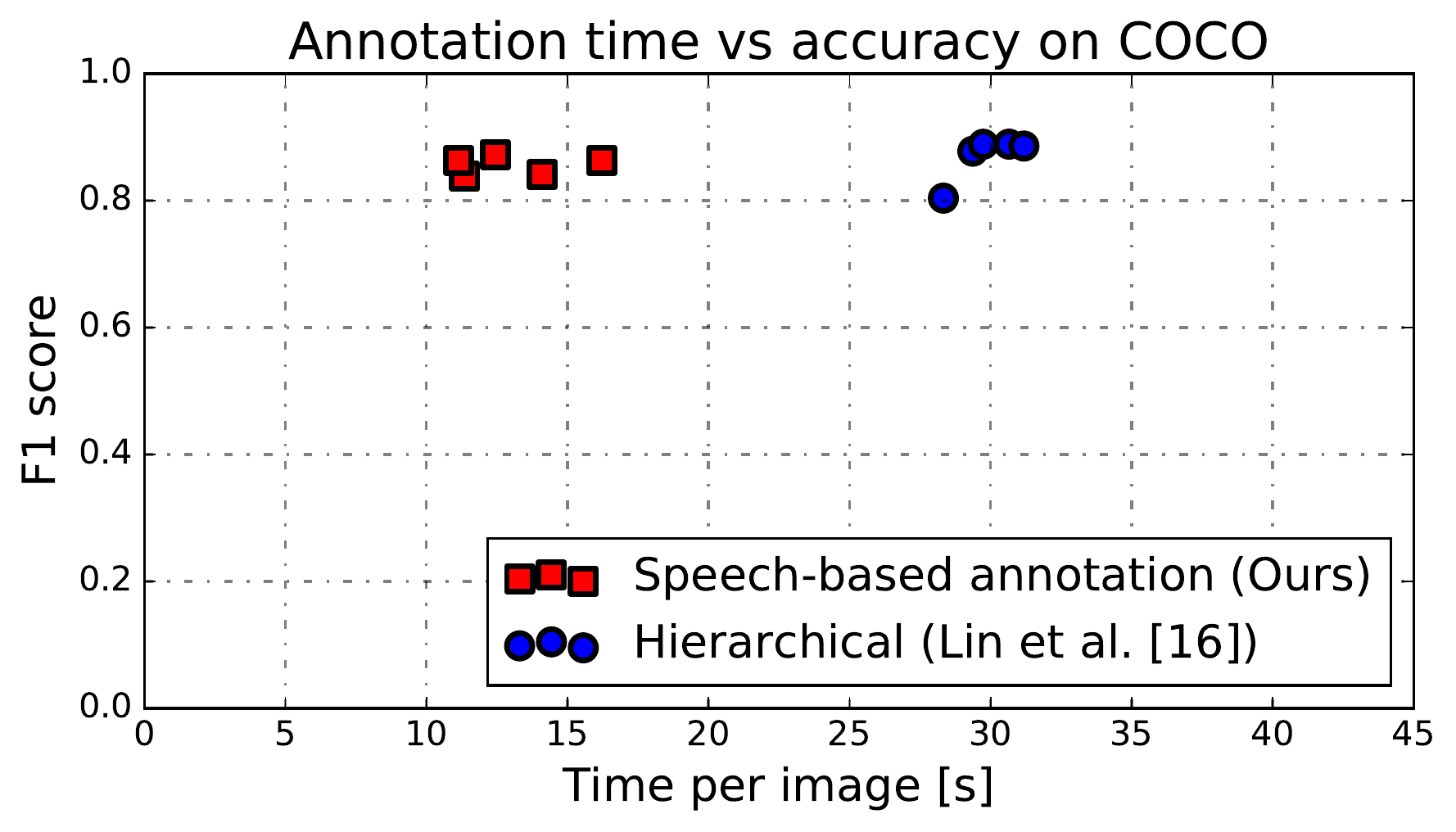}
		\vspace{-0.2cm}
		\caption{Our approach \vs the hierarchical interface of~\cite{lin14eccv}.
		Each point in the plot corresponds to an individual annotator. F1 score is the harmonic mean between recall and precision. Dataset: COCO.}
	\label{fig:results_coco}
\end{figure}

\subsection{Our interface on COCO}
\label{sec:experiments_coco}

In this section we evaluate our approach and compare it to~\cite{lin14eccv}.
Annotations with our interface were done by a new set of crowd workers,
to avoid bias arising from having used the hierarchical interface before.
The workers are all Indian nationals and speak English with an Indian accent.
Hence, we use a model of Indian English for the automatic speech recognition. We also provide the class vocabulary as phrase hints\footnote{\url{https://cloud.google.com/speech-to-text/docs/basics\#phrase-hints}}, which is crucial for obtaining high transcription accuracy of these phrases (Sec.~\ref{sec:detailed_experiments}).

\para{Speed and semantic accuracy.}
Fig.~\ref{fig:results_coco} and Tab.~\ref{tab:accuracy} show results. Our method provides a speed-up of \att{$2.3\times$} over~\cite{lin14eccv} at similar F1 scores (harmonic mean of precision and recall).
In Sec.~\ref{sec:experiments_baseline} we estimated that annotation could be sped up by up to \att{$2.6\times$} by avoiding symbol search and dragging. Interestingly, our interface provides a speedup close to this target, confirming its high efficiency.

Despite the additional challenges of handling speech, average precision is only \att{2\%} lower than for~\cite{lin14eccv}.
Hence, automatic speech transcription does not affect label quality much (we study this further in Sec.~\ref{sec:detailed_experiments}).
Recall is almost identical (\att{0.8\%} lower), confirming that, thanks our training task (Sec.~\ref{sec:training}), annotators remember what classes are in the vocabulary.

\para{Location accuracy.}
We further evaluate the location accuracy of the clicks by using the ground-truth segmentation masks of~\coco{}. Specifically, given an object annotation $o_i$ with class $c_i$, we evaluate whether its click position $p_i$ lies on a ground-truth segment of class $c_i$.
If class $c_i$ is not present in the image at all, we ignore that click in the evaluation to avoid confounding semantic and location errors.

This analysis shows that our interface leads to high location accuracy: \att{$96.0\%$} of the clicks lie on the object.
For the hierarchical interface it is considerably lower at \att{$90.7\%$}.
While this may seems surprising, it can be explained by the differences in the way the location is marked.
In our interface one directly \textit{clicks} on the object, while~\cite{lin14eccv} requires \textit{dragging} a relatively large, semi-transparent class symbol onto it (Fig.~\ref{fig:hierarchical_interface}).

Parts of the speed gains of our interface are due to concurrently providing semantic and location information.
However, this could potentially have a negative effect on click accuracy.
To test this, we compare to the click accuracy that the annotators in~\cite{bearman16eccv} obtained on the PASCAL VOC dataset.
Their clicks have a location accuracy of \att{96.7\%}
comparable to our 96.0\%, despite the simpler dataset with larger objects on average, compared to COCO.
Hence, we can conclude that clicking while speaking does not negatively affect location accuracy.

\subsection{Our interface on \ilsvrc{} 2014}
\label{sec:experiments_imagenet}
Here we apply our interface and the hierarchical interface of~\cite{lin14eccv} to a larger vocabulary of 200 classes, using 300 images from the validation set of \ilsvrc{}~\cite{russakovsky15ijcv}.
For~\cite{lin14eccv} we manually constructed a two-level hierarchy of symbols, based on the multiple hierarchies provided by~\cite{russakovsky15ijcv}.
The hierarchy consists of 23 top-level classes, such as ``fruit'' and ``furniture'', each containing between 5 to 16 object classes.
\begin{table}[t]
	\centering
	\resizebox{1\linewidth}{!}{%
	\setlength{\tabcolsep}{2pt}
	\begin{tabular}{|c|r|r|r|}
		\hline
		&  \textbf{Speech} & \textbf{Lin~\etal~\cite{lin14eccv}} & \textbf{Deng~\etal~\cite{deng14chi}} \\
		\hline
		\multicolumn{4}{|c|}{\textbf{\coco{}}} \\
		Recall     & 83.9 \% & 84.7 \% & \\
		Precision  & 87.3 \% & 89.3 \% & \\

		\hdashline
		Time / image & 13.1s & 29.9s & \\
		Time / label & 4.5s & 11.5s & \\
		\hline
		\multicolumn{4}{|c|}{\textbf{\ilsvrc{}}} \\
		Recall     & 83.4 \% & 88.6 \% & \\
		Precision  & 80.5 \% & 76.6 \% & \\
		\hdashline
		Time / image & 12.0s & 31.1s & $\approx$ 179s~\cite{russakovsky15cvpr} \\
		Time / label & 7.5s & 18.4s & $\approx$ 110s~\cite{russakovsky15cvpr}\\
		\hline
	\end{tabular}
	}
	\caption{Accuracy and speed of our interface (Speech) and hierarchical approaches~\cite{lin14eccv, deng14chi}.
	Our interface is significantly faster at comparable label quality.}
	\label{tab:accuracy}
\end{table}
\begin{figure}[t]
	\centering\includegraphics[width=1\linewidth]{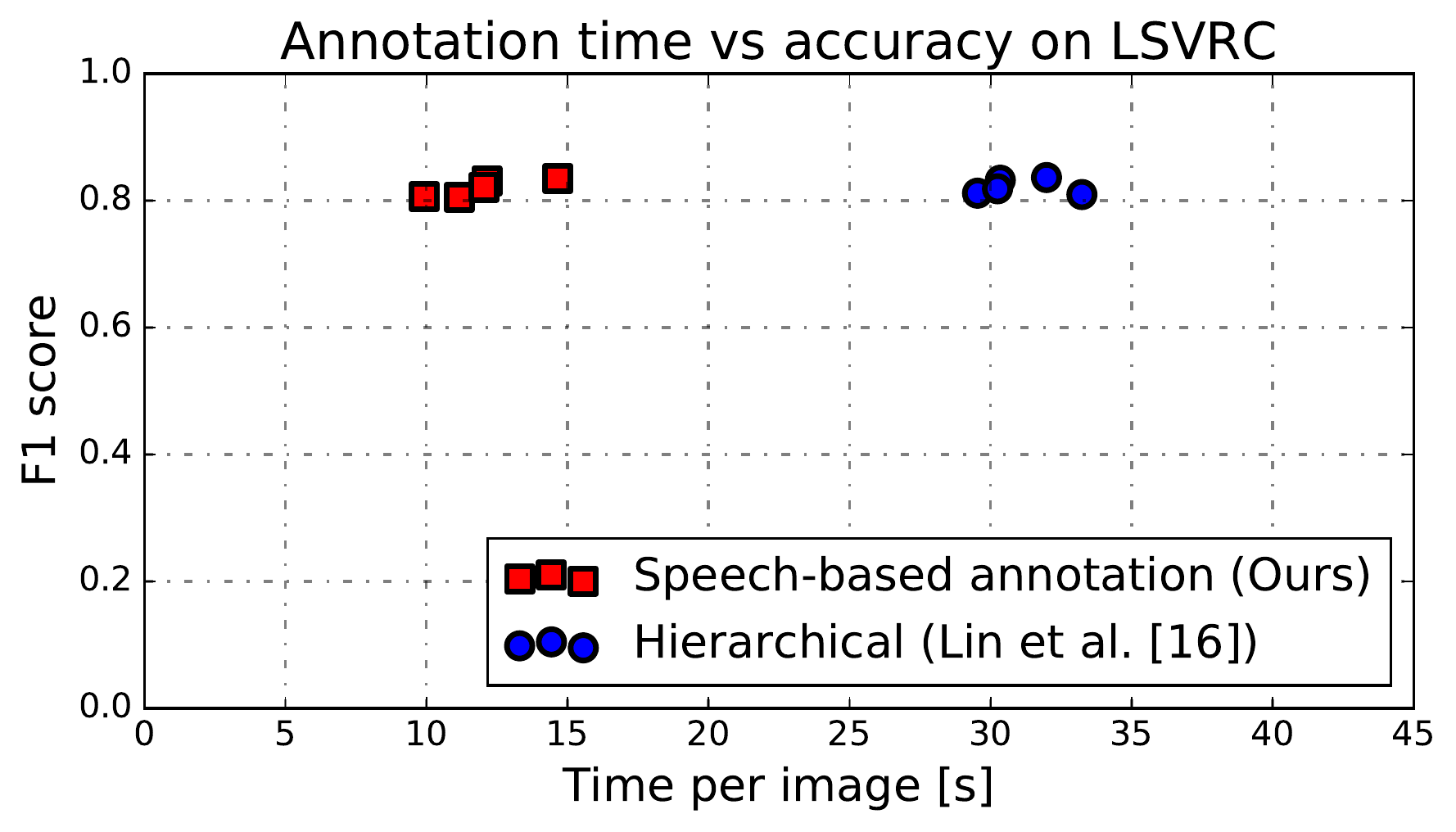}
	\vspace{-0.2cm}
	\caption{Our approach \vs the hierarchical interface~\cite{lin14eccv}. Each point in the plot corresponds to an individual annotator. Dataset: LSVRC.}
	\label{fig:results_imagenet}
\end{figure}

\begin{figure*}[t]
	\centering\includegraphics[width=0.25\linewidth,trim={0 0 0 0},clip]{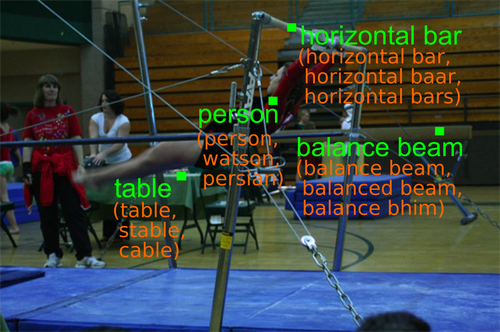}%
	\centering\includegraphics[width=0.25\linewidth,trim={0 0 0 0},clip]{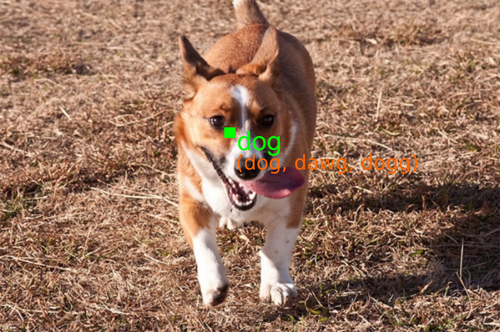}%
	\centering\includegraphics[width=0.25\linewidth,trim={0 1.9cm 0 0},clip]{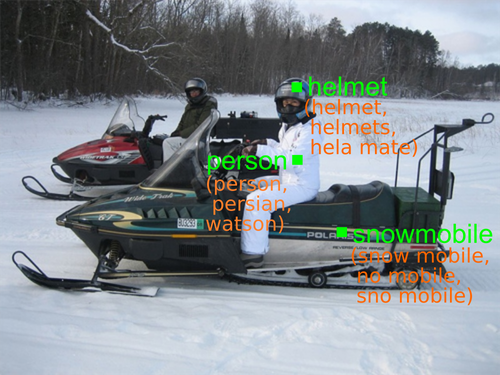}%
	\centering\includegraphics[width=0.25\linewidth,trim={0 0.95cm 0 0},clip]{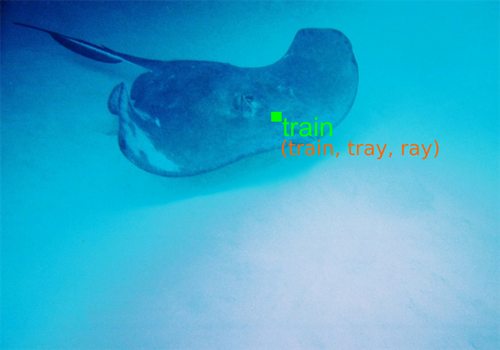}%
	\caption{Example annotations on \ilsvrc{}.
	For each click we show the three alternatives from the ASR model (\textcolor{orange}{orange}) and the final class label (\textcolor{green}{green}).
	The first three images show typical annotations produced by our method.
	The last one shows a failure case:
	while the correct name is among the alternatives, an incorrect transcription matching a class name ranks higher, hence the final class label is wrong.
	}

	\label{fig:examples}
\end{figure*}

\para{Speed and semantic accuracy.}
Fig.~\ref{fig:results_imagenet} shows a comparison to~\cite{lin14eccv} in terms of speed and accuracy, while 
Fig.~\ref{fig:examples} shows example annotations obtained with our interface.
In Tab.~\ref{tab:accuracy}, we also compare to the speed of~\cite{deng14chi}, the method that was used to annotate this dataset. Our approach is substantially faster than both: \att{2.6$\times$} faster than~\cite{lin14eccv} and \att{14.9$\times$} faster than~\cite{deng14chi}.
We also note that~\cite{deng14chi} only produces a list of classes present in an image, while our interface and~\cite{lin14eccv} additionally provide the location of one object per class.

Despite the increased difficulty of annotating this dataset, which has considerably more classes than~\coco{}, our interface produces high-quality labels. The F1 score is similar to that of~\cite{lin14eccv} (\att{81.9\%} \vs \att{82.2\%}). While recall is lower for our interface, precision is higher.

\begin{figure}[t]
	\centering\includegraphics[width=1\linewidth]{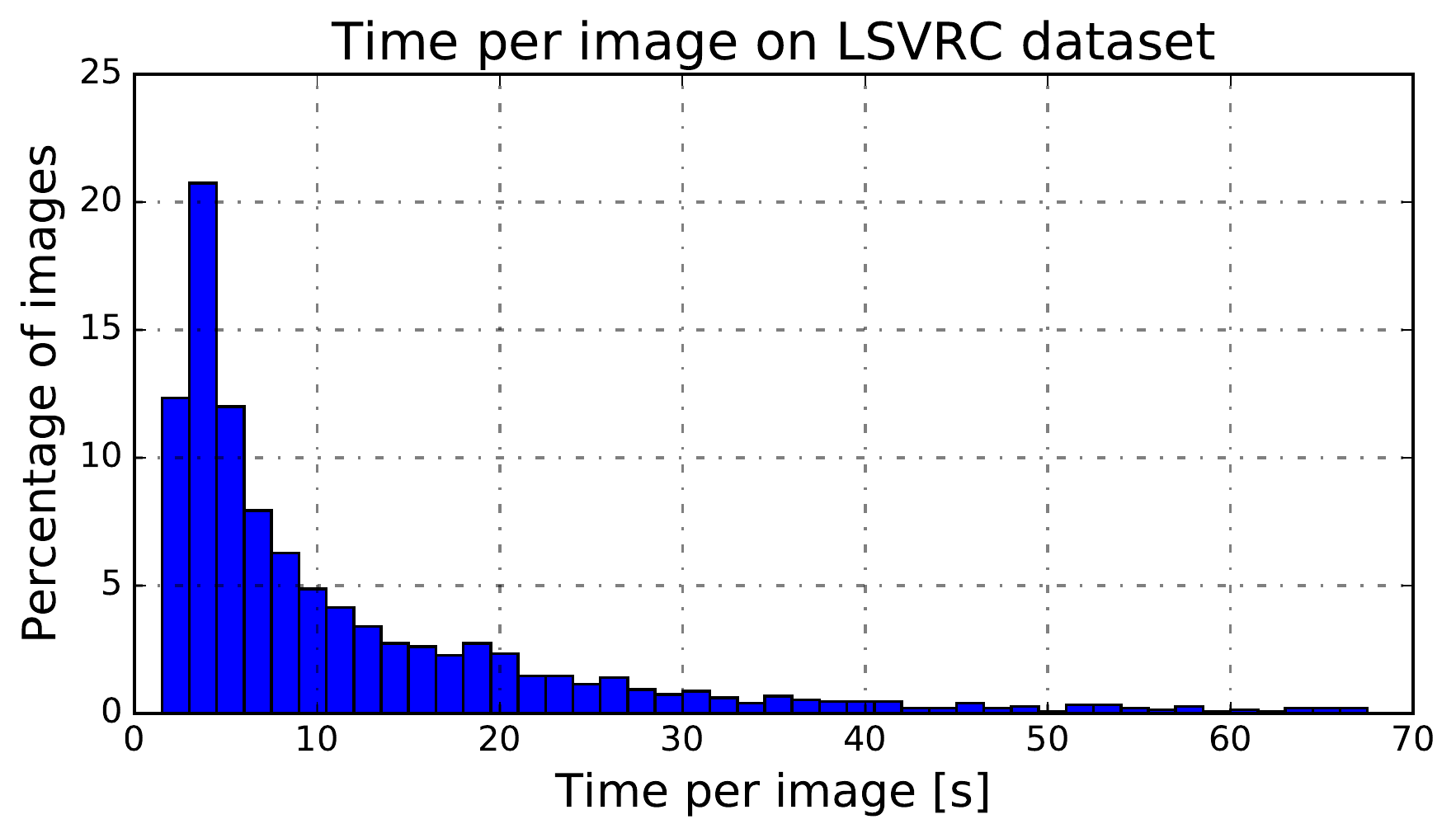}
	\vspace{-0.2cm}
	\caption{Histogram of the time required to annotate an image using our interface. Dataset:~\ilsvrc{}.}
	\label{fig:time_per_image_imagenet}
\end{figure}

Fig.~\ref{fig:time_per_image_imagenet} shows a histogram of the annotation time per image. Most images are annotated extremely fast, despite the large vocabulary, as most images in this dataset contain few classes.
Indeed, there is a strong correlation between the number of object classes present in an image and its annotation time (rank correlation \att{0.55}).
This highlights the advantage of methods that are rooted on the image content, rather than the vocabulary: their annotation time is low for images with few classes. Instead, methods rooted on the vocabulary cannot exploit this class sparsity to a full extent.
The na\"ive approach of asking one yes-no questions per class is actually even slower the fewer objects are present, as determining the absence of a class is slower than confirming its presence~\cite{ehinger09modelling}.

\subsection{Additional analysis of our interface}
\label{sec:detailed_experiments}
\para{Time allocation.}
To understand how much of the annotation time is spent on what, we analyse timings for speaking and moving the mouse on the~\ilsvrc{} dataset.
Of the total annotation time, \att{26.7\%} is spent on speaking.
The mouse is moving \att{74.0\%} of the total annotation time, and \att{62.4\%} of the time during speaking.
The rather high percentage of time the mouse moves during speaking confirms that humans can naturally carry out visual processing and speaking concurrently. %

In order to help annotators label the correct classes, we allowed them to consult the class vocabulary, through a button on the interface (Fig.~\ref{fig:interface}).
This takes \att{7.2\%} of the total annotation time, a rather small share.
Annotators consult the vocabulary in fewer than \att{20\%} of the images. %
When they consulted it, they spent \att{7.8} seconds looking at it, on average.
Overall, this shows the annotators feel confident about the class vocabulary and confirms that our annotator training stage is effective.

\begin{figure}[t]
	\centering\includegraphics[width=1\linewidth]{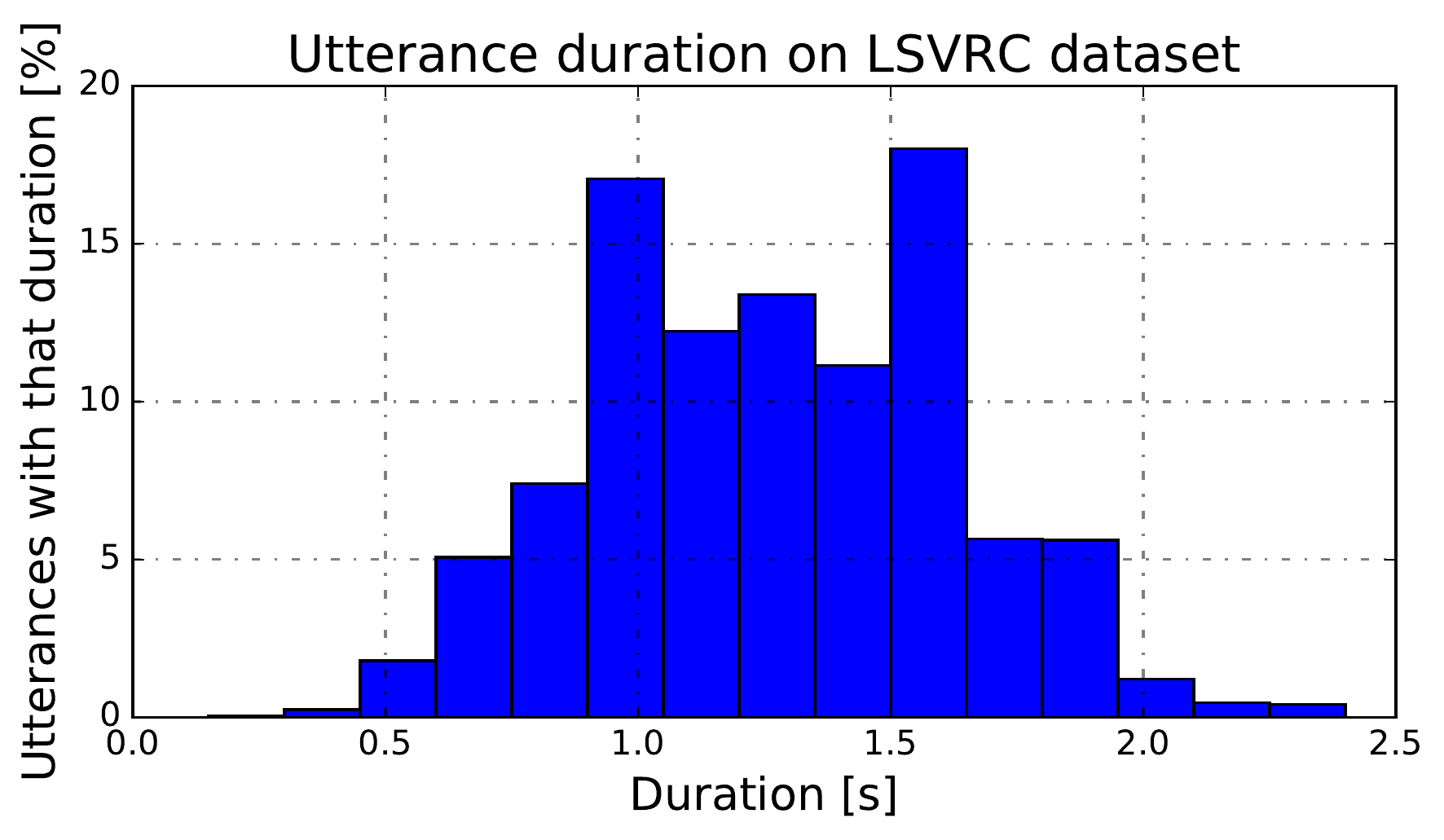}
	\vspace{-0.2cm}
	\caption{Histogram of the time spent \textit{saying} the object name on \ilsvrc{}.
	Saying the object names is fast and usually takes less than 2 seconds.
	}
	\label{fig:utterance_durations}
\end{figure}

In addition, we analyse the time it takes annotators to say an object name in Fig.~\ref{fig:utterance_durations}, which shows a histogram of speech durations. As can be seen, most names are spoken in \att{$0.5$ to $2$ seconds}.

\begin{figure}[t]
	\centering\includegraphics[width=1\linewidth]{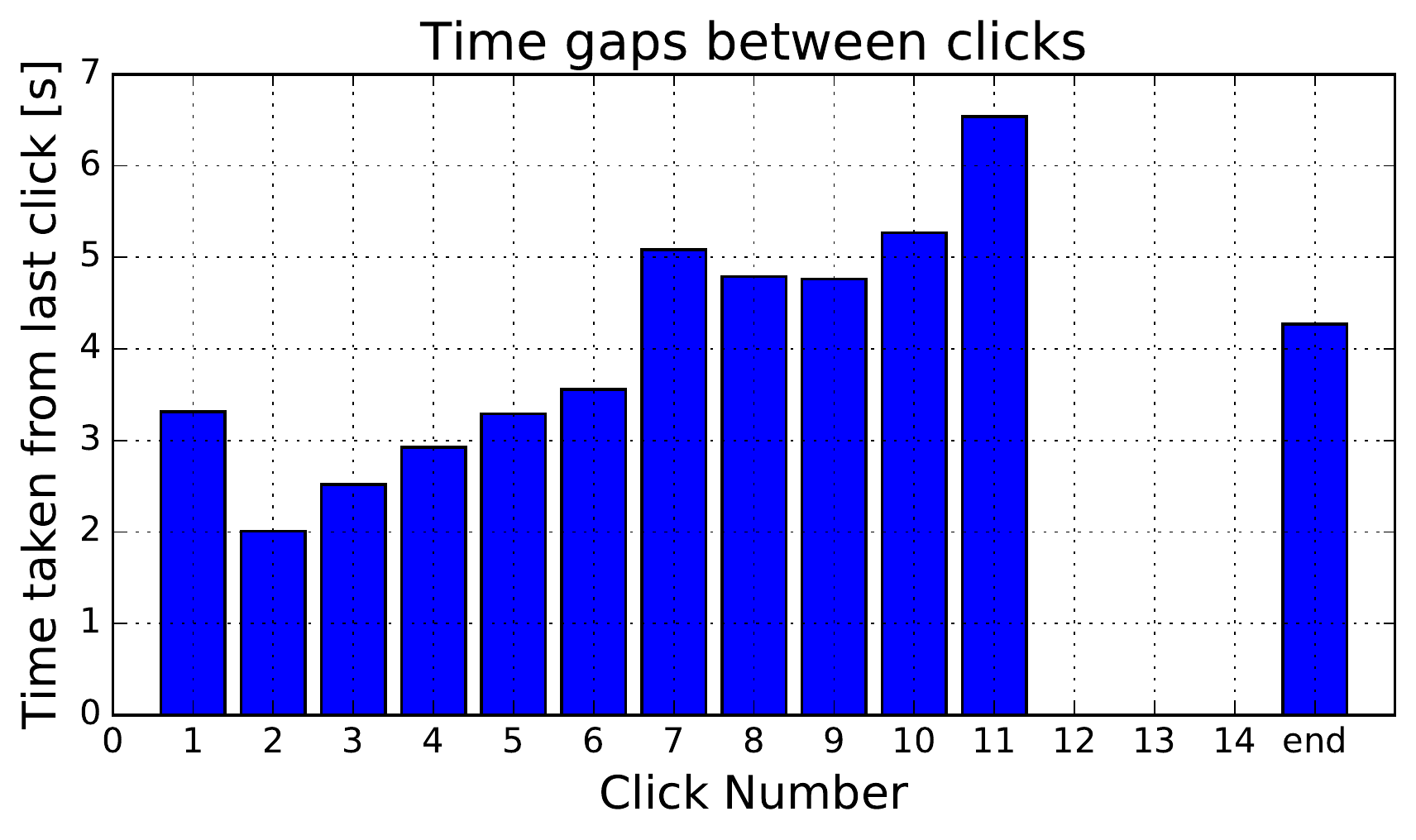}
	\vspace{-0.2cm}
	\caption{Analysis of the time it takes for the first and subsequent clicks when annotating object classes on the \coco{} dataset.}
	\label{fig:time_per_click}
\end{figure}

\para{Per-click response time.}
In Fig.~\ref{fig:time_per_click} we analyse the time taken to annotate the first and subsequent classes of an image in the \coco{} dataset.
It takes \att{3.3}s to make the first click on an object, while the second takes \att{2.0}s only.
This effect was also observed by~\cite{bearman16eccv}.
Clicking on the first object incurs the cost of the initial visual search across the whole scene, while the second is a continuation of this search and thus cheaper~\cite{watson07eye, rayner09, lleras05rapid}.
After the second class, finding more classes becomes increasingly time-consuming again, as large and salient object classes are already annotated.
Indeed, we find that larger objects are typically annotated first: object size has a high median rank correlation with the annotation order (\att{$-0.80$}). Interestingly, on the interface of~\cite{lin14eccv}, this effect is less pronounced (\att{$-0.50$}), as the annotation order is affected by the symbol search and grouping of classes in the hierarchy.
Finally, our analysis shows that the annotators spent \att{3.9}s between saying the last class name and submitting the task, indicating that they do a thorough final scan of the image to ensure they do not miss any class.

\begin{figure}[t]
	\centering
	\includegraphics[trim={0 0 0 1.2cm},clip,width=1\linewidth]{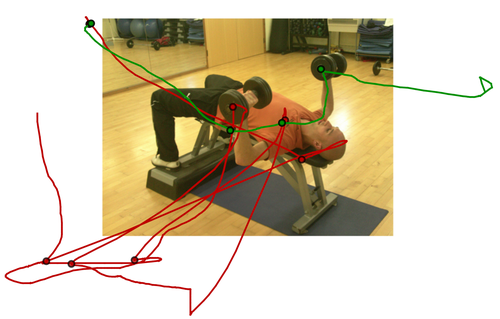}
	\vspace{-0.6cm}
	\caption{A comparison of typical mouse paths produced when annotating an image with our interface (\textcolor{darkgreen}{green}) or with~\cite{lin14eccv} (\textcolor{red}{red}). Circles indicate clicks. Mouse paths for our interface are extremely short, thanks to its simplicity and naturalness.}
	\label{fig:mouse_path}
\end{figure}

\para{Mouse path length.}
To better understand the amount of work required to annotate an image we also analysed the mean length of the mouse path. We find that on \ilsvrc{} annotators using~\cite{lin14eccv} move the mouse for a \att{$3.0\times$} greater length than annotators using our interface.
Thus, our interface is not only faster in terms of time, but is also more efficient in terms of mouse movements.
The reason is that the hierarchical interface requires moving the mouse back and forth between the image and the class hierarchy (Fig.~\ref{fig:mouse_path}).
The shorter mouse path indicates the simplicity and improved ease of use of our interface.

\para{Training time.}
Training annotators to achieve good performance on the 200 classes of~\ilsvrc{} takes 1.6 hours for our interface, or 1 hour with the hierarchical interface of~\cite{lin14eccv}.
Instead, annotating the full~\ilsvrc{} dataset takes 1726 hours with our interface \vs 4474 hours with~\cite{lin14eccv}.
Hence, the cost of training is negligible and our interface is far more efficient than~\cite{lin14eccv} even after
taking training into account.

\para{Transcription accuracy.}
The annotator training task provides spoken and written class names for each annotated object (Sec.~\ref{sec:training}).
Using this data we evaluate the accuracy of the automatic speech recognition (ASR). For this we only take objects into account if they have transcriptions results attached. This keeps the analysis focused on transcription accuracy by ignoring other sources of errors, such as incorrect temporal segmentation or annotators simply forgetting to say the class name
after they click on an object.

Tab.~\ref{tab:transcription_accuracy} shows the transcription accuracy in two setups: with and without using the vocabulary as phrase hints.
Phrase hints allow to indicate phrases or words that are likely to be present in the speech and thus help
the ASR model transcribe them correctly more often.
Using phrase hints is necessary to obtain high transcription accuracy.
Thanks to them, Recall@3 is at \att{96.5\%} on \coco{} and \att{97.5\%} on \ilsvrc{}.
Hence, the top three transcriptions usually contain the correct class name, which we then extract as described in Sec.~\ref{sec:postprocessing}.

In fact, we actually consider above numbers to be a lower bound on the transcription accuracy in the main task, as here we compare the transcriptions against the raw written class names, which contain a few spelling mistakes.
Moreover, here the annotators are in the training phase and hence still learning about the task.
Overall, the above evidence shows that ASR provides high accuracy, definitely good enough for labelling object class names.

\para{Vocabulary usage.}
As speech is naturally free-form, we are interested in knowing how often annotators use object names that are outside of the vocabulary.
Thus, we analyse how often the written class name in the annotator training task does not match a vocabulary name.
We find that on \coco{} annotators are essentially only using names from the vocabulary (\att{99.5\%} of the cases).
On \ilsvrc{} they still mostly use names from the vocabulary, despite the greater number of classes which induces a greater risk of misremembering their names (\att{96.3\%} are in vocabulary).

Some of the out-of-vocabulary names are in fact variations of names in the vocabulary. These cases can be mapped to their correct name in the vocabulary as described in Sec.~\ref{sec:postprocessing}.
For example, for the~\ilsvrc{} dataset some annotators say ``oven'', which gets correctly mapped to ``stove'', and ``traffic signal'' to ``traffic light''.
In other cases the annotators use out-of-vocabulary names because they actually label object classes that are not in the vocabulary (\eg ``fork'' and ``rat'', which are not classes of~\ilsvrc{}).

We find that our annotator training task helps reducing the use of out-of-vocabulary names: on \ilsvrc{} the use of vocabulary names increases from 96.3\% in training to 97.5\% in the main task.

\begin{table}
	\centering
	\resizebox{1\linewidth}{!}{%
	\setlength{\tabcolsep}{12pt}
	\begin{tabular}{|c|r|r|}
		\hline
		&  \textbf{Recall@1} & \textbf{Recall@3} \\
		\hline
		\coco{} w/ hints & 93.1 \% & 96.5 \% \\
		\coco{} w/o hints & 70.5 \% & 84.7 \% \\
		\hdashline
		\ilsvrc{} w/ hints & 93.3 \% & 97.5 \%  \\
		\ilsvrc{} w/o hints & 70.2 \% & 89.5 \%  \\
		\hline
	\end{tabular}
	}
	\caption{\textbf{Transcription accuracy.} Accuracy is high when using phrase hints (see text).
	}
	\label{tab:transcription_accuracy}
\end{table}

\section{Conclusion}

We proposed a novel approach for fast \taskname{}, a task that has traditionally been very time consuming.
At the core of our method lies speech: annotators label images simply by saying the names of the object classes that are present.
In extensive experiments on \coco{} and \ilsvrc{} we have shown the benefits of our method: it offers considerable speed gains of \att{$2.3\times - 14.9\times$} over previous methods~\cite{lin14eccv,deng14chi}.
Finally, we have conducted a detailed analysis of our and previous interfaces, hence providing helpful insights for building efficient annotations tools.

We believe that speech will be useful for other tasks that combine annotating semantic and geometric properties,
because speaking and moving the mouse can naturally be done concurrently~\cite{oviatt03book}.
In fact, our ongoing work shows that when annotating bounding boxes, class labels can be annotated without additional cost.

{\small
\bibliographystyle{ieee}
\bibliography{shortstrings,loco,references}
}

\newpage
\setcounter{section}{0}
\renewcommand{\thesection}{Appendix \Alph{section}}
\section{Two-level Hierarchy for \ilsvrc{}}

For reference we provide the hierarchy we constructed to use the interface of~\cite{lin14eccv} with the 200 class vocabulary of the~\ilsvrc{} dataset~\cite{deng09cvpr}. The hierarchy is based on the hierarchy of questions provided in~\cite{deng09cvpr}, but modified to balance the size of the groups and reduced to two-levels.
It consists of 22 semantic groups and a small group of ``misc objects'':
\small
\begin{enumerate}
\item    Wind instruments:
  \\ \begin{enumerate*}[noitemsep,topsep=0pt,itemjoin={;\quad}]
 \item trumpet
      \item saxophone
      \item trombone
      \item flute
      \item oboe
      \item harmonica
      \item french horn
      \item accordion
   \end{enumerate*}
   
 \item Other musical instruments:
  \\ \begin{enumerate*}[noitemsep,topsep=0pt,itemjoin={;\quad}]
 \item piano
      \item guitar
      \item violin
      \item chime
      \item maraca
      \item drum
      \item cello
      \item banjo
      \item harp
   \end{enumerate*}
   
 \item Fruit:
  \\ \begin{enumerate*}[noitemsep,topsep=0pt,itemjoin={;\quad}]
 \item pineapple
      \item fig
      \item orange
      \item banana
      \item strawberry
      \item apple
      \item lemon
      \item pomegranate
   \end{enumerate*}
   
 \item Other food:
  \\ \begin{enumerate*}[noitemsep,topsep=0pt,itemjoin={;\quad}]
 \item pizza
      \item guacamole
      \item popsicle
      \item hamburger
      \item hotdog
      \item burrito
      \item pretzel
      \item mushroom
      \item bagel
      \item artichoke
      \item cucumber
      \item bell pepper
      \item cabbage
   \end{enumerate*}
   
 \item Clothing:
  \\ \begin{enumerate*}[noitemsep,topsep=0pt,itemjoin={;\quad}]
 \item miniskirt
      \item diaper
      \item brassiere
      \item bathing cap
      \item bow tie
      \item helmet
      \item tie
      \item swimming trunks
      \item swimsuit
      \item hat
      \item sunglasses
   \end{enumerate*}
   
 \item Flying Animals:
  \\ \begin{enumerate*}[noitemsep,topsep=0pt,itemjoin={;\quad}]
 \item bee
      \item ladybug
      \item butterfly
      \item dragonfly
      \item bird
   \end{enumerate*}
   
 \item Felines and Canines:
  \\ \begin{enumerate*}[noitemsep,topsep=0pt,itemjoin={;\quad}]
 \item tiger
      \item lion
      \item domestic cat
      \item fox
      \item dog
   \end{enumerate*}
   
 \item 
    Animals with hooves:
  \\ \begin{enumerate*}[noitemsep,topsep=0pt,itemjoin={;\quad}]
 \item camel
      \item hippopotamus
      \item swine
      \item cattle
      \item zebra
      \item sheep
      \item horse
      \item antelope
   \end{enumerate*}
   
 \item Animals with 6 or more legs:
  \\ \begin{enumerate*}[noitemsep,topsep=0pt,itemjoin={;\quad}]
 \item lobster
      \item scorpion
      \item isopod
      \item centipede
      \item ant
      \item tick
   \end{enumerate*}
   
 \item Animals with no legs:
  \\ \begin{enumerate*}[noitemsep,topsep=0pt,itemjoin={;\quad}]
 \item snake
      \item goldfish
      \item jellyfish
      \item ray
      \item snail
      \item starfish
      \item whale
      \item seal
   \end{enumerate*}
   
 \item Other animals:
  \\ \begin{enumerate*}[noitemsep,topsep=0pt,itemjoin={;\quad}]
 \item red panda
      \item porcupine
      \item giant panda
      \item rabbit
      \item koala
      \item elephant
      \item otter
      \item squirrel
      \item monkey
      \item hamster
      \item skunk
      \item armadillo
      \item bear
      \item frog
      \item lizard
      \item turtle
   \end{enumerate*}
   
 \item Vehicles:
  \\ \begin{enumerate*}[noitemsep,topsep=0pt,itemjoin={;\quad}]
 \item airplane
      \item golfcart
      \item watercraft
      \item train
      \item bus
      \item snowmobile
      \item bicycle
      \item unicycle
      \item snowplow
      \item car
      \item motorcycle
      \item cart
   \end{enumerate*}
   
 \item Cosmetics:
  \\ \begin{enumerate*}[noitemsep,topsep=0pt,itemjoin={;\quad}]
 \item lipstick
      \item face powder
      \item perfume
      \item hair spray
      \item cream
   \end{enumerate*}
   
 \item Medical items:
  \\ \begin{enumerate*}[noitemsep,topsep=0pt,itemjoin={;\quad}]
 \item neck brace
      \item stethoscope
      \item band aid
      \item syringe
      \item stretcher
      \item crutch
   \end{enumerate*}
   
 \item Furniture:
  \\ \begin{enumerate*}[noitemsep,topsep=0pt,itemjoin={;\quad}]
 \item bench
      \item chair
      \item bookshelf
      \item babys bed
      \item table
      \item sofa
      \item filing cabinet
   \end{enumerate*}
   
 \item Carpentry items:
  \\ \begin{enumerate*}[noitemsep,topsep=0pt,itemjoin={;\quad}]
 \item axe
      \item nail
      \item power drill
      \item chain saw
      \item screwdriver
      \item hammer
   \end{enumerate*}
   
 \item School supplies:
  \\ \begin{enumerate*}[noitemsep,topsep=0pt,itemjoin={;\quad}]
 \item pencil box
      \item pencil sharpener
      \item rubber eraser
      \item ruler
      \item binder
   \end{enumerate*}
   
 \item Game equipment:
  \\ \begin{enumerate*}[noitemsep,topsep=0pt,itemjoin={;\quad}]
 \item baseball
      \item golf ball
      \item tennis ball
      \item racket
      \item rugby ball
      \item volleyball
      \item ping-pong ball
      \item croquet ball
      \item basketball
      \item soccer ball
      \item puck
   \end{enumerate*}
   
 \item Sports equipment:
  \\ \begin{enumerate*}[noitemsep,topsep=0pt,itemjoin={;\quad}]
 \item dumbbell
      \item balance beam
      \item horizontal bar
      \item ski
      \item bow
      \item punching bag
   \end{enumerate*}
   
 \item Consumer electronics:
  \\ \begin{enumerate*}[noitemsep,topsep=0pt,itemjoin={;\quad}]
 \item remote
      \item digital clock
      \item computer mouse
      \item computer keypad
      \item laptop
      \item printer
      \item iPod
      \item screen
      \item tape player
      \item microphone
   \end{enumerate*}
   
 \item Electronic appliances:
  \\ \begin{enumerate*}[noitemsep,topsep=0pt,itemjoin={;\quad}]
 \item washer
      \item coffee maker
      \item microwave
      \item waffle iron
      \item toaster
      \item refrigerator
      \item stove
      \item dishwasher
      \item vacuum
      \item electric fan
      \item hair drier
   \end{enumerate*}
   
 \item Non-electric kitchen items:
  \\ \begin{enumerate*}[noitemsep,topsep=0pt,itemjoin={;\quad}]
 \item bowl
      \item ladle
      \item salt shaker
      \item can opener
      \item cocktail shaker
      \item frying pan
      \item spatula
      \item plate rack
      \item strainer
      \item corkscrew
      \item water bottle
      \item mug
      \item pitcher
      \item wine bottle
      \item milk can      
   \end{enumerate*}
   
 \item Misc objects:
  \\ \begin{enumerate*}[noitemsep,topsep=0pt,itemjoin={;\quad}]
 \item person
      \item traffic light
      \item flowerpot
      \item purse
      \item backpack
      \item plastic bag
      \item lamp
      \item beaker
      \item soap dispenser      
   \end{enumerate*}
   \end{enumerate}

\end{document}